# Interpretable Differential Diagnosis with Dual-Inference Large Language Models

Shuang Zhou[1], Mingquan Lin[1], Sirui Ding[2], Jiashuo Wang[3], Genevieve B. Melton[4], James Zou[5], Rui Zhang[1,*]

**Affiliations:**
[1]Division of Computational Health Sciences, Department of Surgery, University of Minnesota, Minneapolis, MN, USA
[2]Bakar Computational Health Sciences Institute, University of California San Francisco, San Francisco, CA, USA
[3]Department of Computing, The Hong Kong Polytechnic University, Hong Kong, Hong Kong SAR
[4]Institute for Health Informatics and Division of Colon and Rectal Surgery, Department of Surgery, University of Minnesota, Minneapolis, MN, USA
[5]Department of Biomedical Data Science, Stanford University, Stanford, CA, USA

*Correspondence: zhan1386@umn.edu

## Abstract

Automatic differential diagnosis (DDx) is an essential medical task that generates a list of potential diseases as differentials based on patient symptom descriptions. In practice, interpreting these differential diagnoses yields significant value but remains under-explored. Given the powerful capabilities of large language models (LLMs), we investigated using LLMs for interpretable DDx. Specifically, we curated the first DDx dataset with expert-derived interpretation on 570 clinical notes. Besides, we proposed Dual-Inf, a novel framework that enabled LLMs to conduct bidirectional inference (i.e., from symptoms to diagnoses and vice versa) for DDx interpretation. Both human and automated evaluation validated its efficacy in predicting and elucidating differentials across four base LLMs. In addition, Dual-Inf could reduce interpretation errors and hold promise for rare disease explanations. To the best of our knowledge, it is the first work that customizes LLMs for DDx explanation and comprehensively evaluates their interpretation performance. Overall, our study bridges a critical gap in DDx interpretation and enhances clinical decision-making.

## Introduction

Differential diagnosis (DDx) is a list of possible conditions that could cause a patient's symptoms [1]. In clinical scenarios, the significance of conducting DDx is multifaceted. Firstly, it provides a more comprehensive assessment of a clinical case than a single diagnosis, facilitating the identification of less apparent yet critical conditions. Secondly, it guides appropriate diagnostic examinations and optimizes resource allocation. Thirdly, by presenting a clear list of potential conditions, clinicians can effectively communicate the diagnostic process and involve patients in decision-making, thereby boosting the credibility of the healthcare system. However, the benefits of DDx come at the cost of the additional time and effort required from clinicians, which overwhelms the overburdened workforce.

Several automatic DDx algorithms [2, 3] have been developed to alleviate clinician workload. These approaches typically utilized deep learning (DL) techniques for supervised learning to provide DDx predictions. For example, Liu et. al [4] developed a DL-based system to distinguish among 26 common skin conditions. Hwang et. al [5] proposed a convolutional neural network-long short-term memory (CNN-LSTM) model for cardiovascular diseases. Additionally, another DL-based platform was designed to identify 39 types of referable fundus diseases [6]. Nevertheless, merely providing diagnostic predictions to clinicians is still insufficient as the black-box nature of DL models often undermines trust [7]. To enhance their reliability, it is essential to provide interpretative insights into the diagnoses [8]. Consequently, interpretable DDx is highly desirable in clinical scenarios. This task takes patients' symptom descriptions as input, generates a set of potential diseases as differentials, and provides explanations for these diagnoses. Despite its importance, interpretable DDx remains under-explored due to various challenges. Firstly, there is a shortage of DDx datasets with interpretation of diagnoses for model development and evaluation [9, 10]. Secondly, the complexity of diagnosing diseases and delivering coherent explanations typically requires large models with extensive medical knowledge, which entails significant costs in data collection and model training.

Large language models (LLMs), such as ChatGPT, trained on millions of corpora (including medical data), have exhibited remarkable capabilities across a wide range of natural language processing (NLP) tasks [11-14]. Numerous studies have highlighted their effectiveness in various clinical scenarios, including medical question answering (QA) [15-17], clinical text summarization [18], and disease diagnosis [19-24]. For instance, Med-PaLM 2 [25], an instruction-tuned LLM with over 340 billion parameters, achieved comparable performance with clinicians on medical QA datasets. Similarly, DRG-LLaMA [26], fine-tuned on 236,192 MIMIC-IV discharge summaries using the LLaMA model [27], achieved an AUC score of approximately 0.99 for Diagnosis-Related Group (DRG) prediction. Motivated by these advancements, employing LLMs for interpretable DDx appears promising.

Although recent studies have adopted LLMs for DDx, they primarily focus on diagnostic accuracy while falling short of examining interpretation. For instance, Daniel et al. [28] fine-tuned PaLM 2 on medical data and developed an interactive interface to assist clinicians with DDx generation. Tu et al. [29] built an LLM-based system for clinical history-taking and diagnostic dialogue by simulating patient interactions. Although Savage et al. [30] have employed Chain-of-thought (CoT) [31] to harness LLMs' reasoning ability for boosting diagnostic accuracy, how to effectively leverage LLMs for DDx explanation and thoroughly evaluate their interpretation performance remains under-explored.

In this work, we formally investigated prompting LLMs for interpretable DDx without pretraining or fine-tuning on massive data. Our key contributions are as follows. Firstly, we curated a new dataset comprising 570 clinical notes across nine specialties. Specifically, the data was derived from publicly available medical corpora, such as books (e.g., *First Aid for the USMLE Step 2 CS* [32]), and was annotated by domain experts with a set of differentials and the corresponding interpretations. To the best of our knowledge, it is the first publicly available structured dataset with DDx interpretation annotation [9, 33]. Secondly, we proposed a prompt-based framework called Dual-Inf to effectively elicit LLMs' capabilities for explanation generation. The key design is to enable LLMs to perform bidirectional inference (i.e., from symptoms to diagnoses and vice versa) and leverage the backward verification to examine and boost the predictions' correctness. Specifically, Dual-Inf consists of (1) a forward-inference module for generating initial diagnoses based on patient notes, (2) a backward-inference module that objectively recalls representative symptoms from the initial diagnoses, (3) an examination module, which receives patients' notes and the outputs from the other two modules for correctness examination and decision making, and (4) an iterative self-reflection mechanism, which iteratively takes low-confidence diagnoses as feedback to elicit correct predictions. Thirdly, we comprehensively evaluated Dual-Inf for interpretable DDx, including model interpretability, robustness on rare diseases, and error analysis. The results demonstrated that Dual-Inf not only achieved superior diagnostic performance but also provided faithful interpretations across various base LLMs (i.e., GPT-4, GPT-4o, Llama3-70B, and BioLlama3-70B). Additionally, Dual-Inf made fewer errors than the baselines, such as SC-CoT [34] and Self-Contrast [35]. Overall, we verified the efficacy of Dual-Inf for interpretable DDx, making it a promising tool for enhancing clinical decision-making, and the annotated DDx dataset could advance the development of the research field.

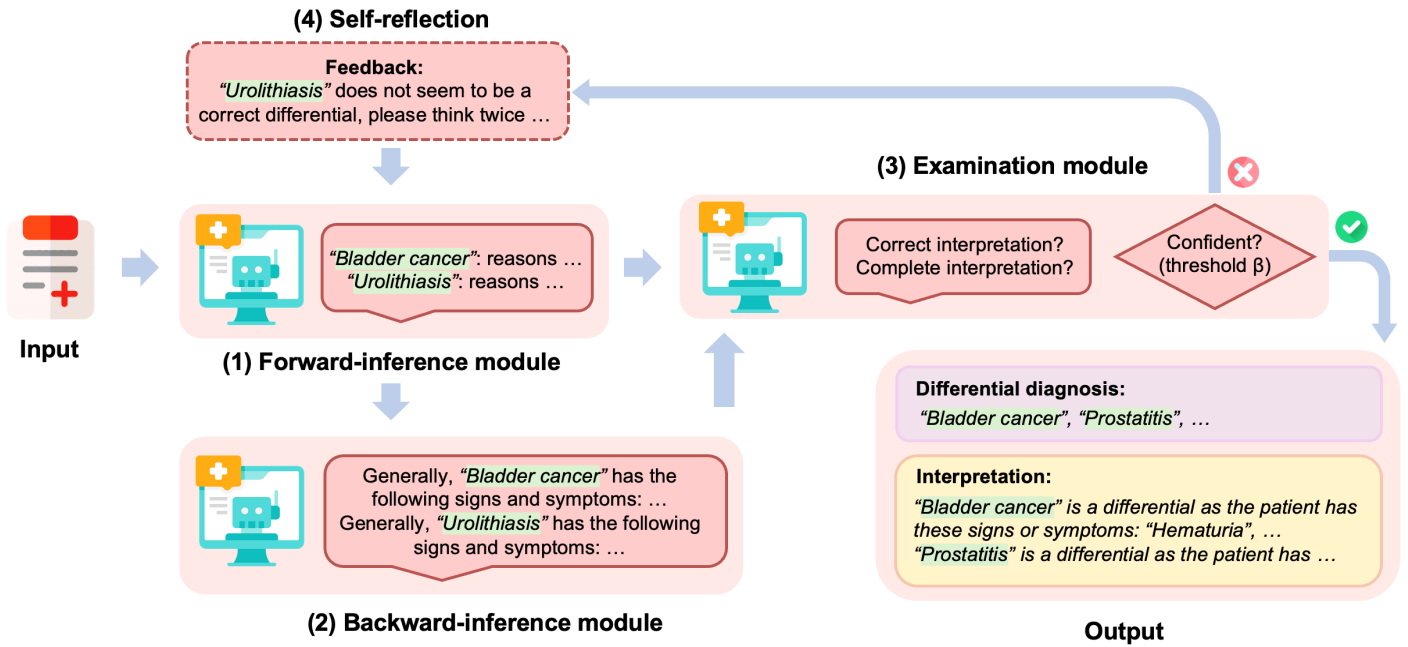

Figure 1. An overview of the Dual-Inference Large Language Model framework (Dual-Inf) for interpretable DDx. Dual-Inf consists of four components: (1) a forward-inference module, which is an LLM to generate initial diagnoses from patient symptoms, (2) a backward-inference module, which is an LLM for conducting inverse inference via recalling all the representative symptoms associated with the initial diagnoses, i.e., from diagnoses to symptoms, (3) an examination module, which is another LLM to receive patients' notes and the output from the two modules for prediction assessment (e.g., completeness examination) and decision making (e.g., filtering out low-confidence diagnoses), and (4) an iterative self-reflection mechanism, which iteratively takes the low-confidence diagnoses as feedback for the forward-inference module to "think twice".

## Results

### Dataset

We created Open-XDDx, a well-annotated text-based dataset for interpretable DDx, comprising 570 clinical notes from publicly available medical exercises. This dataset spans 9 clinical specialties: cardiovascular disease, digestive system disease, respiratory disease, endocrine disorders, nervous system disease, reproductive system disease, circulatory system disease, skin disease, and orthopedic disease. Each note was well-organized and involved diverse information, including patient symptom descriptions, differential diagnoses, and the corresponding interpretations supporting the potential diagnoses. Since the original data lacked interpretations, clinicians from the University of Minnesota annotated the notes. The annotation procedure was described in Supplementary Data 1. The statistics of the dataset are shown in Table 1.

Table 1. The data characteristics of our annotated interpretable DDx dataset Open-XDDx.

| Statistic | Number | Clinical Specialty | Number |
|---|---|---|---|
| # notes | 570 | Cardiovascular disease | 26 (4.6%) |
| note length (mean) | 113.6 | Digestive system disease | 105 (18.4%) |
| note length (std) | 60.4 | Respiratory disease | 58 (10.2%) |
| # diagnosis (mean) | 4.6 | Endocrine disorder | 43 (7.5%) |
| # diagnosis (std) | 1.0 | Nervous system disease | 137 (24.0%) |
| # reasons per patient (mean) | 14.5 | Reproductive system disease | 54 (9.5%) |
| # reasons per patient (std) | 5.8 | Circulatory system disease | 66 (11.6%) |
| # reasons per diagnosis (mean) | 3.1 | Skin disease | 30 (5.3%) |
| # reasons per diagnosis (std) | 1.5 | Orthopedic disease | 51 (8.9%) |

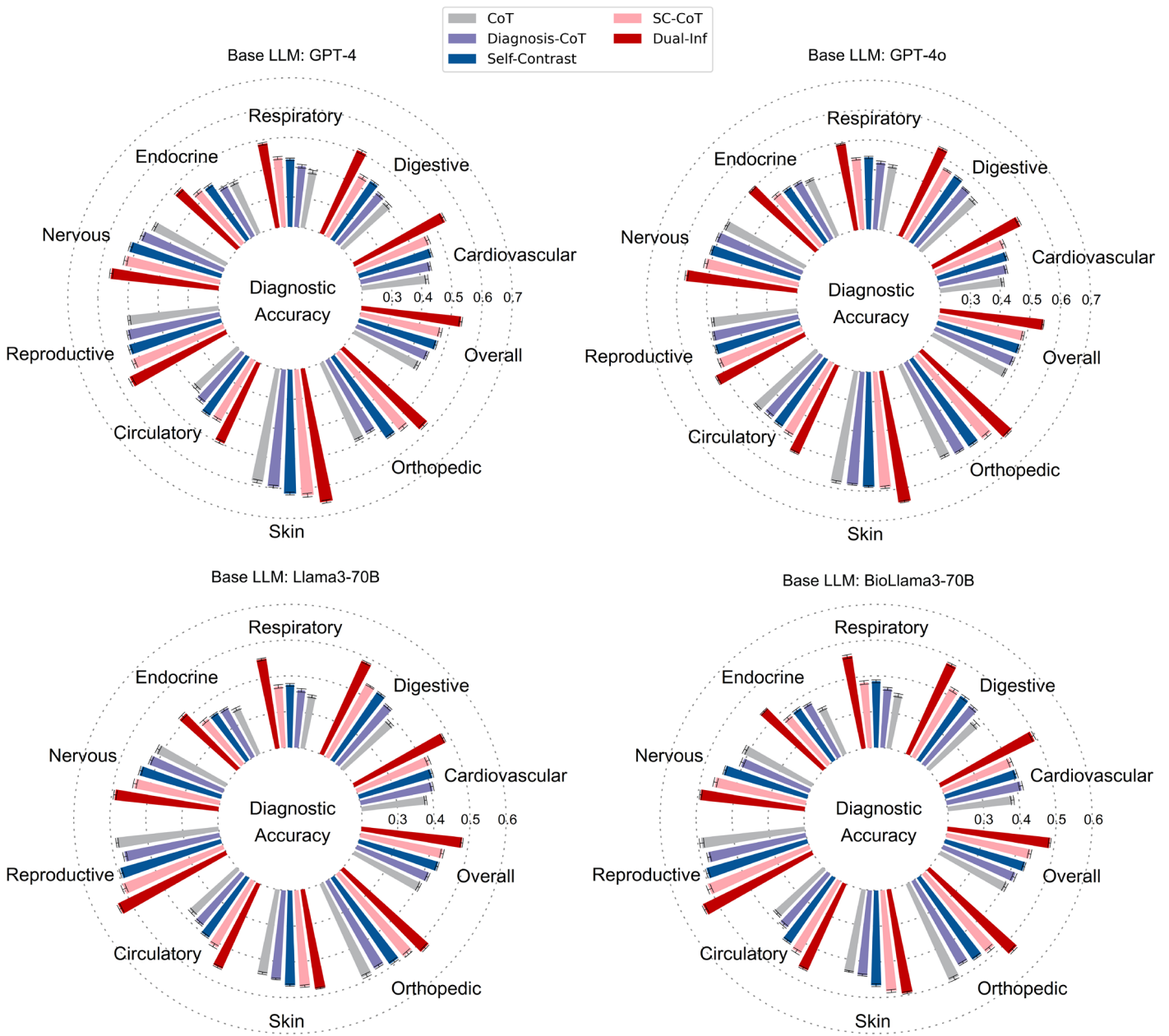

Figure 2. Differential diagnosis performance based on four base LLMs (GPT-4, GPT-4o, Llama3-70B, and BioLlama3-70B) over nine specialties. The results are averaged over five runs. Standard deviations are also shown.

Differential Diagnosis performance

We evaluated differential diagnosis accuracy (Eq. 1) through automatic comparison of ground-truth diagnoses with model predictions. The prompts are shown in Supplementary Data 2. The results with GPT-4, GPT-4o, Llama3-70B, and BioLlama3-70B are depicted in Figure 2. It showed that Dual-Inf consistently outperformed the baselines across the nine specialties. Specifically, with GPT-4 as the base LLM, the overall performance of SC-CoT significantly exceeded CoT (difference of 0.032, 95% CI 0.021 to 0.043, $p$=0.001) and Diagnosis-CoT (difference of 0.019, 95% CI 0.001 to 0.028, $p$=0.004). Dual-Inf further surpassed SC-CoT (0.533 vs. 0.472, difference of 0.061, 95% CI 0.055 to 0.062, $p$=<0.001). Similarly, using GPT-4o, Dual-Inf achieved over 0.55 accuracy on nervous, skin, and orthopedic diseases, exceeding SC-CoT by over 9%. When taking Llama3-70B and BioLlama3-70B as the base LLM, Dual-Inf outperformed SC-CoT by over 10% in cardiovascular, digestive, and respiratory diseases. The overall performance improvement of Dual-Inf over SC-CoT across the three base LLMs (difference of 0.059, 0.048, and 0.049) was statistically significant ($p$=<0.001).

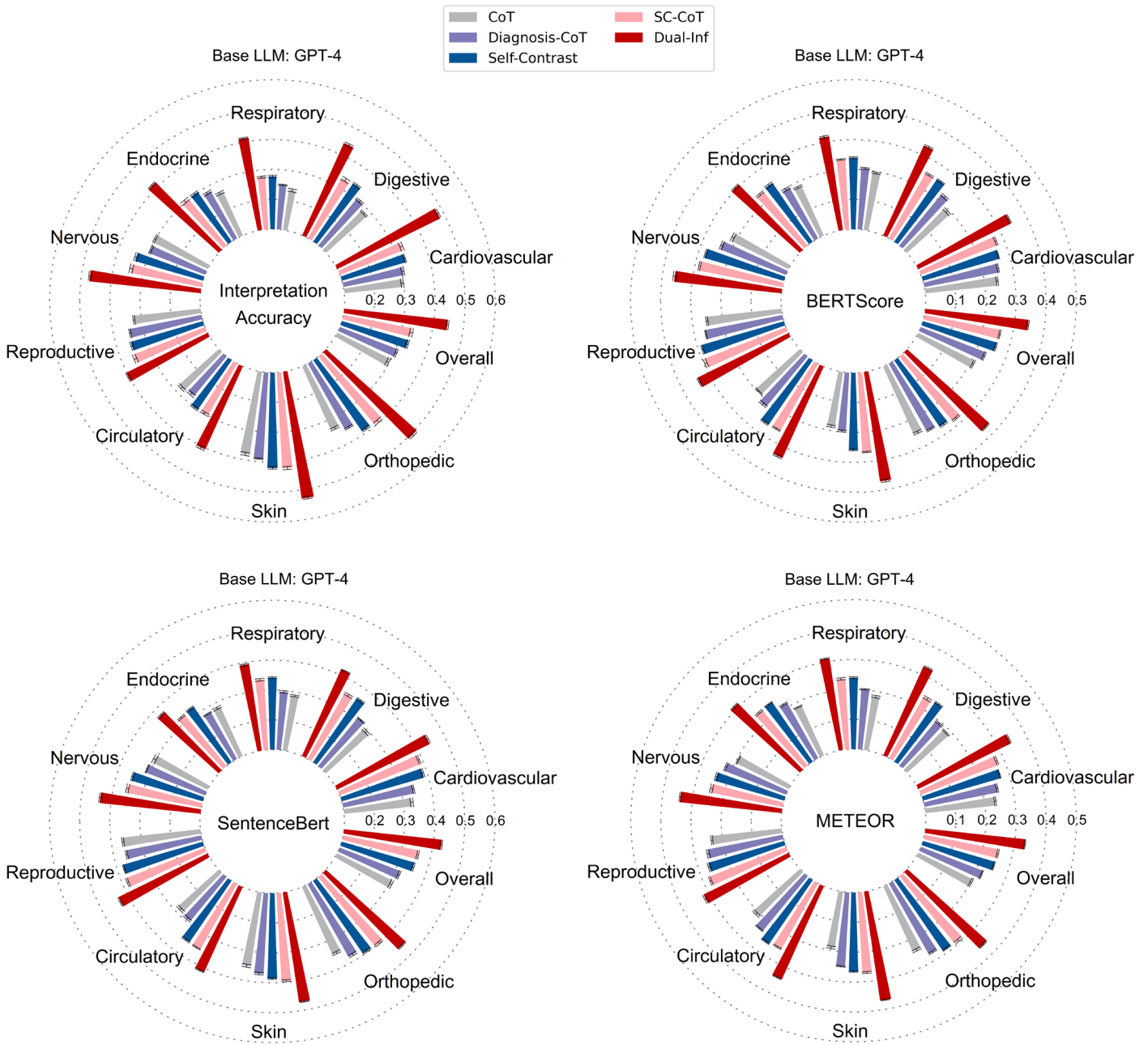

Figure 3. Interpretation performance w.r.t four metrics, i.e., interpretation accuracy (see Eq. 2), BERTScore, SentenceBert, and METEOR across nine clinical specialties. We implemented the methods with GPT-4. The results were averaged over five runs. Standard deviations were also shown.

Interpretation performance

We assessed model interpretability using both automatic and human evaluations. Regarding automatic evaluation, we used GPT-4o to compare the consistency between ground-truth interpretations with the predicted ones. The prompts are detailed in Supplementary Data 2. We tested four base LLMs (GPT-4, GPT-4o, Llama3-70B, and BioLlama3-70B). The results for GPT-4 are shown in Figure 3, with additional results in Supplementary Data 3. In Figure 3, the interpretation accuracy (Eq. 2) of Diagnosis-CoT and SC-CoT was 0.305 and 0.334, surpassing CoT by 0.011 (95% CI 0.004 to 0.019, $p$=0.012) and 0.04 (95% CI 0.038 to 0.043, $p$=<0.001), respectively. Dual-Inf achieved even higher accuracy at 0.446, with a 0.112 improvement over SC-CoT (95% CI 0.105 to 0.118, $p$=<0.001). For BERTScore, SentenceBert, and METEOR, Dual-Inf also outperformed SC-CoT with comparisons of 0.345 vs. 0.258 (difference of 0.087, 95% CI 0.083 to 0.090, $p$=<0.001), 0.427 vs. 0.356 (difference of 0.071, 95% CI 0.067 to 0.076) and 0.333 vs. 0.251 (difference of 0.082, 95% CI 0.076 to 0.088). When taking GPT-4o as the base LLM, the interpretation accuracy of Dual-Inf reached 0.488, outperforming CoT and SC-CoT, which scored 0.366 and 0.408, respectively. On the other metrics, Dual-Inf consistently surpassed SC-CoT, with differences of 0.083, 0.064, and 0.08. Similarly, with Llama3-70B and BioLlama3-70B, Dual-Inf exceeded SC-CoT by over 17% across all metrics.

The interpretations were also manually examined by clinicians based on correctness, completeness, and usefulness. Figure 4(a) presents the results for 100 randomly selected notes, using GPT-4 as the base LLM. We observed that the correctness scores of Dual-Inf predominantly ranged from 3 to 4, whereas SC-CoT scores mainly fell between 2 and 3. In terms of interpretation completeness, Dual-Inf achieved 38 scores of 3 and 21 scores of 4, compared to SC-CoT's 19 and 3, respectively. Regarding overall usefulness, Dual-Inf had 33 scores of 3 and 25 scores of 4, while SC-CoT had 26 and 10, respectively.

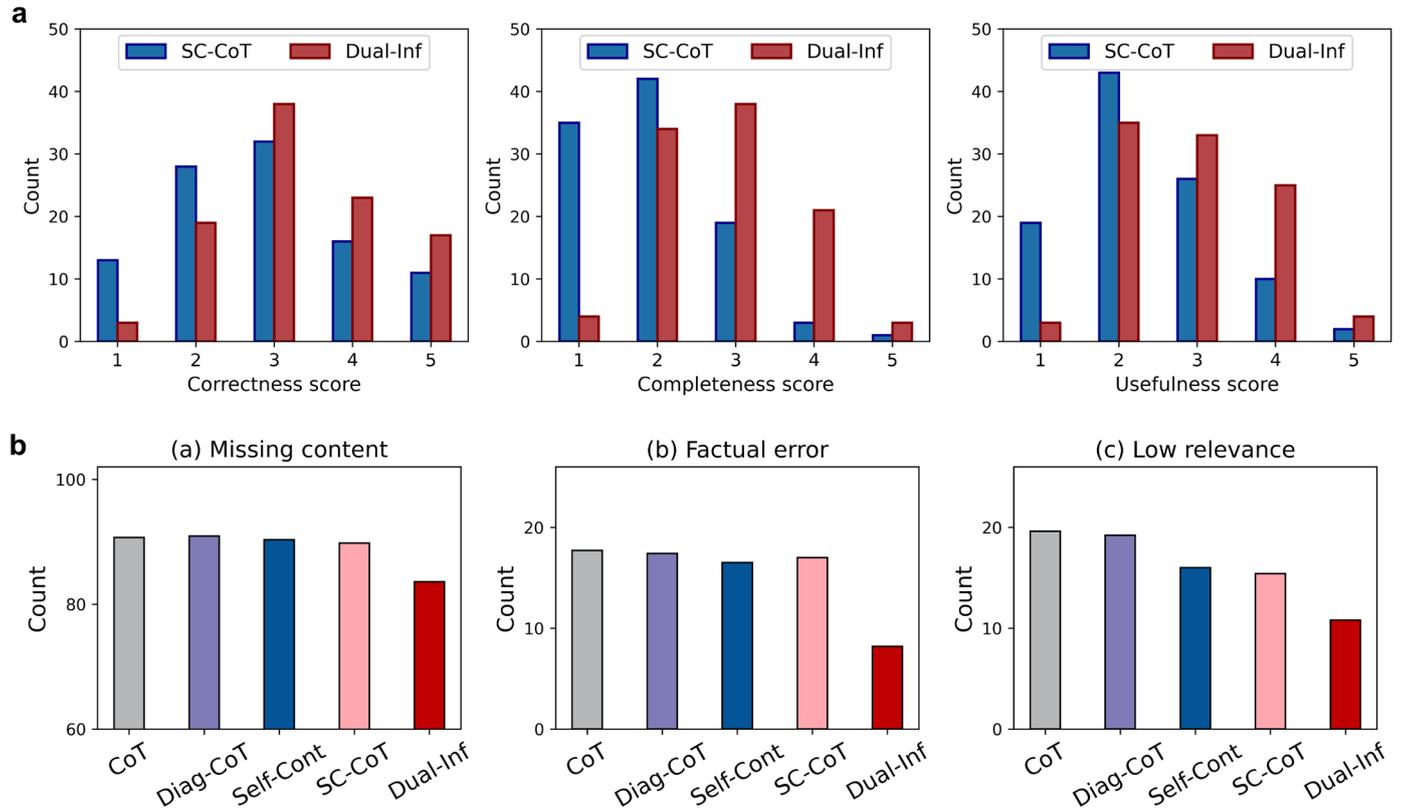

Figure 4. (a) Human evaluation results on interpretation. It assessed three aspects: correctness, completeness, and usefulness with scores ranging from 1 to 5. (b) Error type analysis on interpretation. We manually examined 100 cases and recorded the count of the error type. Diag-CoT denotes Diagnosis-CoT and Self-Cont means Self-Contrast. The results were averaged over five runs. The methods are implemented with GPT-4.

### Error analysis on interpretation

We further analyzed the error types in the generated interpretations. Specifically, we randomly selected 100 samples where SC-CoT produced incorrect interpretations and compared them to Dual-Inf. Following related studies [36, 37], errors were categorized into three categories: (1) missing content, defined as missing at least two pieces of evidence for a differential; (2) factual error, where the interpretation is medically incorrect; (3) low relevance, indicating that the evidence provided was not highly pertinent to the differential. As shown in Figure 4(b), when using GPT-4 as the base LLM, SC-CoT had over 89 predictions with missing content, while Dual-Inf had only about 76 (difference of 13.4, 95% CI 11.5 to 15.2). For factual error and low relevance, the count number comparison between Dual-Inf and SC-CoT was 17 vs. 8.2 (difference of 8.8, 95% CI 7.8 to 9.8) and 15.4 vs. 10.8 (difference of 4.6, 95% CI 3.9 to 5.3), respectively. These differences were statistically significant ($p$=<0.001).

### Performance on rare disease

We further investigated the performance of Dual-Inf on the rare disease cases in the dataset. Table 2 presents the results using GPT-4 as the base LLM, and the results with GPT-4o can be found in Supplementary Table 5. In terms of diagnostic accuracy, the comparison between CoT and Diagnosis-CoT was 0.396 vs. 0.402

(difference 0.006, 95% CI -0.001 to +0.012, *p*=0.073), and the score difference between CoT and SC-CoT was 0.029 (95% CI 0.017 to 0.040, *p*=0.002). The DDx accuracy of Dual-Inf was 0.483, exceeding SC-CoT by (0.058) 13%. As for the interpretation metrics, Dual-Inf scored 0.379, 0.284, 0.371, and 0.275, while those of SC-CoT were 0.285, 0.207, 0.316, and 0.220. When taking GPT-4o as the base LLM, we observed similar results. Concretely, Dual-Inf outperformed SC-CoT in both DDx accuracy (0.491 vs. 0.431, difference 0.060, 95% CI 0.052 to 0.068, *p*=<0.001) and interpretation metrics (0.426 vs. 0.354, 0.287 vs. 0.211, 0.378 vs. 0.326, and 0.286 vs. 0.235).

Table 2. Interpretable DDx performance (and standard deviation) with GPT-4 on rare diseases.

| Method | Diagnostic Performance | Interpretation Performance | | | |
|---|---|---|---|---|---|
| | Accuracy | Accuracy | BERTScore | SentenceBert | METEOR |
| CoT | 0.396 ± 0.006 | 0.253 ± 0.004 | 0.179 ± 0.004 | 0.264 ± 0.003 | 0.169 ± 0.002 |
| Diagnosis-CoT | 0.402 ± 0.005 | 0.259 ± 0.005 | 0.185 ± 0.002 | 0.272 ± 0.002 | 0.186 ± 0.003 |
| Self-Contrast | 0.417 ± 0.004 | 0.278 ± 0.004 | 0.198 ± 0.002 | 0.304 ± 0.003 | 0.212 ± 0.002 |
| SC-CoT | 0.425 ± 0.004 | 0.285 ± 0.005 | 0.207 ± 0.003 | 0.316 ± 0.003 | 0.220 ± 0.003 |
| Dual-Inf | **0.483 ± 0.005** | **0.379 ± 0.004** | **0.284 ± 0.002** | **0.371 ± 0.002** | **0.275 ± 0.002** |

## Ablation study

We further conducted an ablation study to analyze the impact of each component within Dual-Inf. Specifically, we built four method variants for comparison: (1) a method using only the forward-inference module (FI), (2) Dual-Inf without the backward-inference module (FI-EM), (3) FI-EM without the self-reflection mechanism (FI-EM*), and (4) Dual-Inf without the self-reflection mechanism (Dual-Inf*). The results using GPT-4 as the base LLM are shown in Table 3 and the results based on GPT-4o are presented in Supplementary Table 6. In Table 3, we observed that the diagnostic accuracy of FI, FI-EM*, and Dual-Inf* was below 0.46, while FI-EM approached 0.5. Dual-Inf further improved diagnostic accuracy by approximately 7% compared to FI-EM. As for interpretability, FI and FI-EM* performed similarly across the four metrics. The interpretation accuracy of Dual-Inf* and FI-EM was around 0.358 and 0.373, respectively. Besides, Dual-Inf presented superior interpretability, with BERTScore boosting by 24% and METEOR by 28% over the Dual-Inf*.

Table 3. Ablation study results with GPT-4.

| Method | Diagnostic Performance | Interpretation Performance | | | |
|---|---|---|---|---|---|
| | Accuracy | Accuracy | BERTScore | SentenceBert | METEOR |
| FI | 0.443 ± 0.008 | 0.294 ± 0.006 | 0.209 ± 0.004 | 0.307 ± 0.004 | 0.202 ± 0.003 |
| FI-EM* | 0.449 ± 0.006 | 0.302 ± 0.005 | 0.219 ± 0.002 | 0.320 ± 0.002 | 0.216 ± 0.002 |
| FI-EM | 0.497 ± 0.007 | 0.373 ± 0.007 | 0.265 ± 0.004 | 0.374 ± 0.003 | 0.286 ± 0.003 |
| Dual-Inf* | 0.453 ± 0.004 | 0.358 ± 0.003 | 0.276 ± 0.003 | 0.356 ± 0.003 | 0.261 ± 0.003 |
| Dual-Inf | **0.533 ± 0.005** | **0.446 ± 0.004** | **0.343 ± 0.003** | **0.427 ± 0.002** | **0.335 ± 0.002** |

## Discussion

The first observation of our study was that Dual-Inf boosted DDx accuracy. This was because Dual-Inf filtered out low-confidence diagnoses by examining the quality of interpretation. Specifically, the examination module integrated outputs from the other two modules to verify correctness, while the self-reflection mechanism allowed the forward-inference module to iteratively receive feedback and refine its predictions. To assess the impact of the iterative reflection, we tracked the number of iterations for each note in Dual-Inf, as shown in Figure 5(a). The results revealed that most predictions were revised iteratively. Besides, we selected ten notes with five iterations and recorded the number of correct diagnoses after each iteration in Figure 5(c). It showed that Dual-Inf tended to generate more accurate diagnoses after iterations, verifying the effectiveness of iterative reflection in Dual-Inf. We further visualized the distribution of the DDx accuracy with GPT-4 on each case in Figure 5(b). The median and three-quarters percentile of diagnostic accuracy with Dual-Inf were 0.495 and 0.746, respectively, outperforming SC-CoT (0.434 and 0.652) and Diagnosis-CoT (0.421 and 0.628). These improvements were statistically significant ($p=<0.001$), verifying Dual-Inf's effectiveness in boosting diagnostic accuracy across most notes.

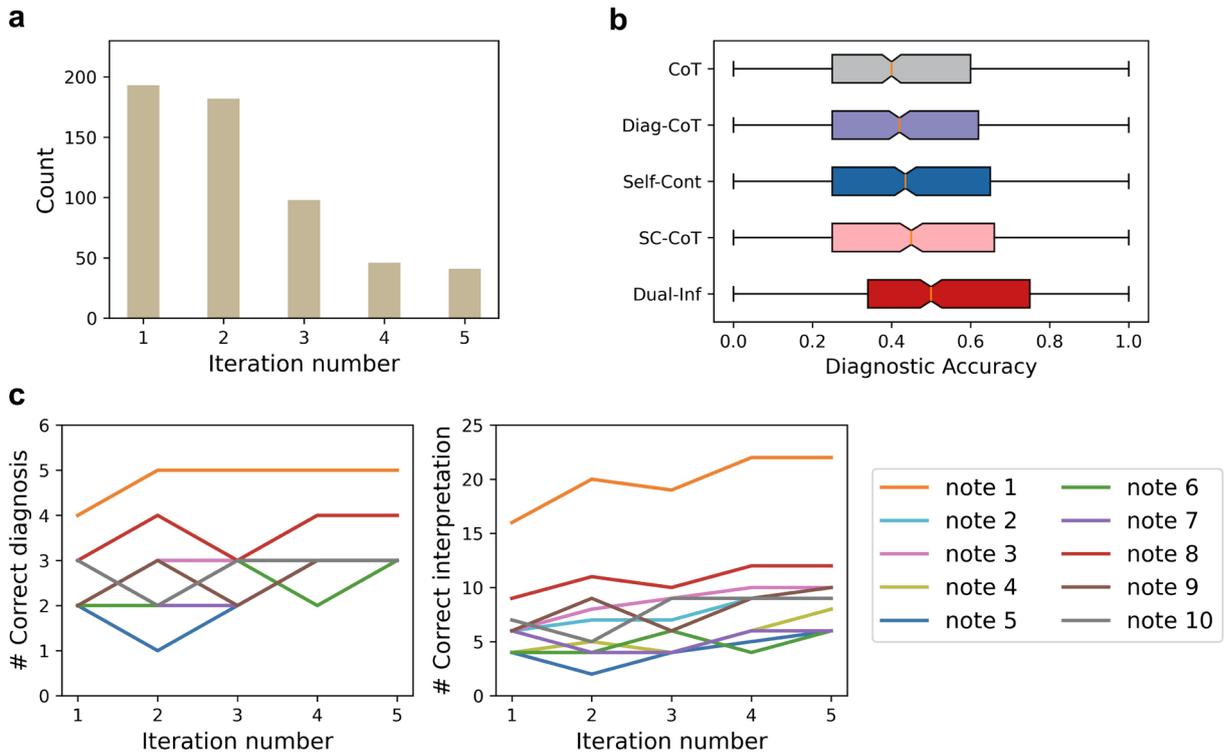

Figure 5. (a) Data statistics of the iteration number for each note in Dual-Inf. (b) Distribution visualization of diagnostic accuracy on each note. Diag-CoT denotes Diagnosis-CoT and Self-Cont means Self-Contrast. (c) Performance change on Dual-Inf w.r.t diagnosis and interpretation after each iteration. We randomly selected ten notes with five iterations. In this figure, the base LLM of the methods is GPT-4.

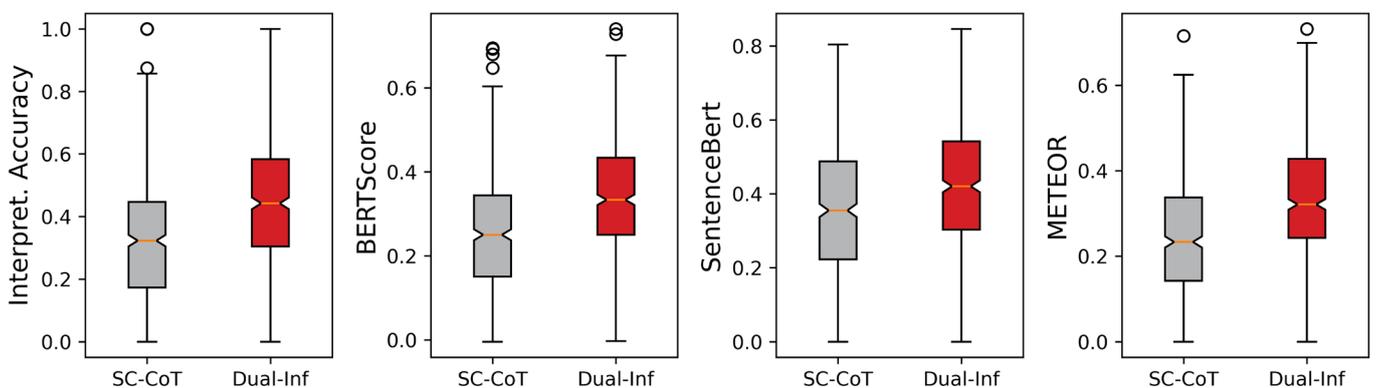

Figure 6. Distribution visualization of interpretation performance on each note. SC-CoT and Dual-Inf were implemented with GPT-4. The circular points shown as outliers mean that some scores are deviated from the vast majority.

Second, we found that Dual-Inf provided better interpretations for DDx. For example, the manual examination of 100 cases (Figure 4(b)) showed that Dual-Inf achieved higher scores across all the evaluation metrics. The superiority lies in that Dual-Inf made bidirectional inferences to supplement the interpretations and more importantly, it iteratively revised the prediction. To verify the effectiveness of iterative reflection, we selected ten notes with five iterations and reported the number of correct interpretations after each iteration in Figure 5(c). Besides, we used a case study showing the intermediate prediction results after each iteration in Supplementary Data 4. The results showed that Dual-Inf generally obtained better interpretations after the iterations. Additionally, we provided a distribution visualization of the interpretation on each note in Figure 6; the methods were implemented with GPT-4. We observed that the median and quarter percentiles of the metric scores (e.g., BERTScore and METEOR) on Dual-Inf were larger than the baseline method SC-CoT. It revealed that Dual-Inf generated better interpretations for most of the cases.

We further provided another case study to demonstrate the superior interpretability of Dual-Inf over the baselines. The example in Table 4 showcased that Diagnosis-CoT and SC-CoT only provided three correct interpretations for a differential, i.e., *Pneumothorax*, while Dual-Inf generated more accurate interpretations. This was because the bi-directional inference enabled recalling all the related symptoms from diagnoses, thus providing comprehensive explanations. Besides, Dual-Inf had one more correct differential, i.e., *Hemothorax*, with three correct interpretations than the baselines. The reasons were as follows: (1) Dual-Inf could filter out the low-confidence diagnosis, i.e., *Pulmonary Contusion*, which had a limited number of interpretations as evidence support, (2) the self-reflection mechanism provided another chance for prediction and took the low-confidence diagnosis as feedback to elicit a correct answer. See more examples in Supplementary Data 5.

Table 4. Case study of Diagnosis-CoT, SC-CoT, and Dual-Inf. The methods are implemented by taking GPT-4 as the base LLM. Correct predictions are highlighted in blue.

| |
|---|
| **Patient's symptom description:** |
| 25-year male complains of left chest pain and LUQ pain following an MVA. The patient struck a tree with his car at a slow speed. The chest pain is 8110. It is exacerbated with movement or when he takes a deep breath, and nothing relieves it. He reports dyspnea and a productive cough with a low-grade fever but denies LOC, headache, change in mental status, or change in vision. No cardiovascular or neurologic symptoms. No nausea, vomiting, neck stiffness, or unusual fluid from the mouth or nose. No dysuria. His last meal was 5 hours ago. He denies being under the influence of alcohol or drugs. ROS: As per HPI. Allergies: NKDA. Medications: None. PMH: Infectious mononucleosis 2 months ago. PSH: None. SH: No smoking, occasional EtOH, no illicit drugs. FH: Noncontributory. Physical Examination Patient is in acute distress, dyspneic. VS: Temp 100°F, RR 22/minute. HEENT: Atraumatic, no JVD, no bruises, PERRLA, EOMI, no pharyngeal edema or exudates. Chest: Two large ecchymosis on left chest, left rib tenderness, decreased breath sounds over left lung field, right lung fields clear. Heart: RRR; S1/S2 WNL; no murmurs, rubs, or gallops. Abdomen: Soft, nondistended, BS x 4 quadrants, LUQ tenderness, no rebound or guarding, no organomegaly. Skin: No bruises or lacerations. Neuro: Mental status: Alert and oriented x 3. Cranial nerves: 2-12 grossly intact. Motor: Strength 515 in all muscle groups. Sensation: Intact to pinprick and soft touch. |
| **Ground-truth DDx:** |
| *Pneumothorax, Hemothorax, Pneumonia* |
| **Ground-truth interpretation:** |
| The patient may suffer from *Pneumothorax* because of the following symptoms or evidence: |
| "Left-sided chest pain following an MVA", "Decreased breath sounds over left lung field", "Pain is exacerbated by movement and deep breaths", "RR 22/minute", "Dyspnea", "Bruising and tenderness over left chest" |
| The patient may suffer from *Hemothorax* because of the following symptoms or evidence: |
| "Left-sided chest pain following an MVA", "Decreased breath sounds over left lung field", "Dyspnea", "RR 22/minute", "Cough", "Bruising and tenderness over left chest" |
| The patient may suffer from *Pneumonia* because of the following symptoms or evidence: |
| "Unilateral chest pain", "Temperature 100°F", "Productive cough", "RR 22/minute", "Low-grade fever" |
| **Diagnosis-CoT (GPT-4) - predicted DDx:** |
| *Rib Fracture, Pneumothorax, Pulmonary Contusion* |
| **Diagnosis-CoT (GPT-4) - predicted interpretation:** |
| The patient may suffer from *Rib Fracture* because of the following symptoms or evidence: "Left chest pain and LUQ pain following an MVA", "Two large ecchymosis on left chest", "Left rib tenderness" |
| The patient may suffer from *Pneumothorax* because of the following symptoms or evidence: "Decreased breath sounds over left lung field", "Dyspnea", "Chest pain exacerbated with movement or deep breath" |
| The patient may suffer from *Pulmonary contusion* because of the following symptoms or evidence: "Productive cough with a low-grade fever", "Decreased breath sounds over left lung field" |
| **SC-CoT (GPT-4) - predicted DDx:** |
| *Rib Fracture, Pneumothorax, Pulmonary Contusion* |
| **SC-CoT (GPT-4) - predicted interpretation:** |
| The patient may suffer from *Rib Fracture* because of the following symptoms or evidence: "Left chest pain and LUQ pain following an MVA", "Two large ecchymosis on left chest", "Left rib tenderness", "Chest pain is exacerbated with movement or when taking a deep breath" |
| The patient may suffer from *Pneumothorax* because of the following symptoms or evidence: "Dyspnea", "Decreased breath sounds over left lung field", "Chest pain exacerbated with movement or when taking a deep breath" |
| The patient may suffer from *Pulmonary Contusion* because of the following symptoms or evidence: "Decreased breath sounds over left lung field", "Productive cough with a low-grade fever" |
| **Dual-Inf (GPT-4) - predicted DDx:** |
| *Rib Fracture, Pneumothorax, Hemothorax* |
| **Dual-Inf (GPT-4) - predicted interpretation:** |
| The patient may suffer from *Rib Fracture* because of the following symptoms or evidence: "Left chest pain and LUQ pain following an MVA", "Two large ecchymosis on left chest", "Left rib tenderness", "Chest pain is exacerbated with movement or deep breath" |
| The patient may suffer from *Pneumothorax* because of the following symptoms or evidence: "Dyspnea", "Decreased breath sounds over left lung field", "Chest pain exacerbated with movement or deep breath", "RR 22/minute" |
| The patient may suffer from *Hemothorax* because of the following symptoms or evidence: "Left-sided chest pain following an MVA", "Decreased breath sounds over left lung field", "Dyspnea" |

Third, this study verified that using multiple LLMs could mitigate interpretation errors in DDx. As illustrated in Figure 4(b), SC-CoT produced fewer low-relevance interpretation errors than CoT. This indicated incorporating the interpretations from multiple LLMs helped to alleviate hallucinations. Compared to the baselines, Dual-Inf made fewer erroneous interpretations across the three error types. We attributed it to the dual-inference scheme. Concretely, the forward-inference module generated initial diagnoses from symptoms, while the backward-inference module objectively recalled medical knowledge (i.e., from diagnoses to symptoms), the examination module then refined the interpretations and filtered out low-confidence predictions, and the self-reflection mechanism further provided another chance to "think twice". These designs collectively reduced hallucinations and boosted the quality of DDx interpretation.

Finally, our study demonstrated a promising tendency to use Dual-Inf for rare disease diagnosis. Generally, patients with rare diseases have a higher probability of underdiagnosis or misdiagnosis [38, 39]. Although recent works investigated using LLMs for rare disease diagnosis [40-42], the feasibility of prompting LLMs to provide interpretation on rare diseases was under-explored. To the best of our knowledge, our work is the first one to explore the problem. We believe the explanation for DDx can boost the trustworthiness of the prediction and assist doctors in identifying less obvious but critical diagnoses for patients with rare diseases. We found that SC-CoT, which leveraged multiple LLMs, outperformed CoT on disease diagnosis, and Dual-

Inf further outperformed SC-CoT (Table 2). As for interpretation, Dual-Inf brought about a 10% performance improvement over SC-CoT on BERTScore and METEOR. The above results indicated that GPT-4 had the medical knowledge on the rare disease and Dual-Inf could effectively elicit the diagnostic ability of LLMs, e.g., GPT-4. Considering that Dual-Inf generalizes well to arbitrary LLMs, it is feasible to adopt advanced LLMs in the future, thus being promising for rare disease diagnosis and interpretation.

One limitation of this study is that our dataset, encompassing nine clinical specialties, does not fully capture the breadth of real-world scenarios. Additionally, the dataset lacks annotations on the priority of each diagnosis within the DDx, as ranking the likelihood of possible diseases presents significant challenges.

In summary, this study established a manually annotated dataset for interpretable DDx and introduced a prompt-based framework Dual-Inf that enabled multiple LLMs to perform bi-directional inference (i.e., from diagnoses to symptoms and vice versa). The findings revealed that existing prompting methods exhibited suboptimal performance in generating DDx and interpretations, limiting their practical utility in clinical scenarios. Our experiments verified the effectiveness of Dual-Inf for providing accurate DDx, delivering comprehensive interpretations, and reducing prediction errors. Besides, Dual-Inf holds promise for rare disease diagnosis and explanation. The above results underscored the potential of LLMs in advancing interpretable DDx. Furthermore, the released dataset with ground-truth DDx and interpretations could facilitate the research field. Future work could expand the dataset to a broader range of clinical specialties or integrate domain knowledge from external databases for superior performance.

# Methods

## Data acquisition and processing

The data source for this study is publicly available medical exercises collected from MedQA USMLE (United States Medical Licensing Exam) dataset [43] and professional medical books, including *First Aid for the USMLE Step 2 CS* [32] and *First Aid Q&A for the USMLE Step 1* [44]. The raw data was multiple-choice question answering exercises. We transformed the questions into free text by preserving the patients' symptom descriptions and removing the multiple-choice options. The texts were further preprocessed including (1) removing duplicate notes, (2) unifying all characters into UTF-8 encoding and removing illegal UTF-8 strings; (3) correcting or removing special characters; (4) filtering out notes with fewer than 130 characters. The full dataset can be found in Supplementary Data 6.

## Data annotation

The data source was multiple-choice medical exercise, and the raw data generally did not have the annotation of DDx, interpretation, and clinical specialty. To build a well-annotated dataset for evaluation, we employed three clinical physicians to curate the dataset manually. Two independent physicians annotated each exercise. When disagreement existed in the annotation, a third physician examined the case and made the final annotation. We checked the inter-annotator agreement (IAA) on DDx, interpretation, and specialty. Concretely, we used the Jaccard score [45] to measure the IAA for DDx and interpretation. The value of the Jaccard score ranged from 0 to 1; a larger score denotes better agreement. The results showed that the IAA scores on DDx and interpretation were 0.885 and 0.859, reflecting a great agreement between the annotators. As for the specialty, we adopted Cohen's Kappa score (κ) [46] for IAA evaluation. In detail, we first considered one of the annotations as ground truth and the other as predictions then rotated it and took the average score as the result. The κ score on specialty was 0.914, showing a high consistency in annotation. See more details in Supplementary Data 1. Additionally, the annotated dataset is well-structured in a standardized format, facilitating automated evaluation.

## Model development

Although LLMs exhibit superior performance in various clinical scenarios, they present limited reasoning ability and struggle to handle complicated tasks such as disease diagnosis. To alleviate the issue, a previous study adopted CoT and its variants to perform diagnostic inference step-by-step. However, it might achieve suboptimal performance for DDx explanation, a reasoning-intensive task involving multiple diagnoses and requiring a holistic view for inference. Therefore, how to effectively elicit LLMs' capability for accurate DDx interpretation is a non-trivial problem. Inspired by the phenomenon that humans usually conduct backward reasoning to validate the correctness of answers when solving reasoning problems, we propose performing backward verification (i.e., from diagnosis to symptoms) to examine the predicted diagnoses and eliciting correct answers via self-reflection.

As shown in Figure 1, we developed a framework called Dual-Inference Large Language Model (Dual-Inf) for interpretable DDx. Its input was clinical notes, and the output involved a set of potential diagnoses and the reasons that supported the prediction. Specifically, Dual-Inf consisted of four components: (1) a forward-inference module, which was an LLM for initial diagnoses, i.e., from patients' symptoms to diagnoses, (2) a backward-inference module, which was an LLM for inverse inference via recalling all the representative symptoms of the initial diagnoses, i.e., from diagnoses to symptoms, (3) an examination module, which was another LLM that received patients' notes and the output from the two modules for prediction assessment and decision making, and (4) an iterative self-reflection mechanism, which iteratively took low-confidence diagnoses as feedback for the forward-inference module to "think twice".

The pipeline is as follows. First, the forward-inference module analyzed the clinical notes to infer initial diagnoses and provide interpretations. Next, the backward-inference module conducted backward inference, i.e., from diagnoses to symptoms. It received the initial diagnoses as input and recalled the associated symptoms, including medical examination and laboratory test results. Afterward, the examination module processed all the information for verification and decision-making. Specifically, (i) it checked the correctness of the interpretation from the forward-inference module and discarded erroneous ones by comparing with the recalled knowledge, (ii) it supplemented the interpretations by integrating patient notes with recalled knowledge, (iii) it made the final decisions on whether to accept or filter out predictions based on their quality. The underlying idea was that a diagnosis supported by fewer interpretations was deemed less trustworthy. To assess the confidence of a diagnosis, a threshold $\beta$ was applied; if the number of supporting interpretations was below $\beta$, the diagnosis was regarded as low-confident. Later, the self-reflection mechanism took the low-confidence diagnoses as feedback to prompt the forward-inference module to "think twice". This process iterates, with a maximum iteration limit $\lambda$ set to balance effectiveness and efficiency. Upon reaching this limit, the framework outputs the results. In this study, we implemented the three modules with publicly accessible LLMs, i.e., GPT-4, GPT-4o, Llama3-70B, and BioLlama3-70B, and evaluated their performance on our dataset. The prompts for the three modules are detailed in Supplementary Data 7.

Implementation details

We adopted four baselines: CoT [31], Diagnosis-CoT [30], Self-Contrast [35], and Self-consistency CoT (SC-CoT) [34]. In our implementation, SC-CoT generated five reasoning paths for each note and then selected the most consistent diagnoses and interpretations. The prompt of the baselines can be found in Supplementary Data 8. As for Dual-Inf, the maximum iteration number $\lambda$ was assigned to 5, the threshold $\beta$ was set to 3. For a fair comparison, the methods were implemented with the same base LLM. The base LLMs include GPT-4, GPT-4o, Llama3-70B[1], and BioLlama3-70B[2]. For the former two LLMs, we used the API from the OpenAI company[3], which were "gpt-4-turbo-preview" and "gpt-4o"; for the latter two, we downloaded the models from Huggingface for inference. The temperature parameter was set as 0.1. We further analyzed the impact of the hyper-parameter $\beta$ on the performance and presented the results in Supplementary Data 9.

---

[1] https://huggingface.co/meta-llama/Meta-Llama-3-70B
[2] https://huggingface.co/aaditya/Llama3-OpenBioLLM-70B
[3] https://platform.openai.com/docs/models

## Automatic evaluation metric

Considering that our established dataset contains ground-truth labels, we conduct automatic evaluation by comparing the ground-truth annotation with the predicted ones. Following related papers [47], we used accuracy as the primary metric for assessing diagnostic performance, i.e.,

$$\text{Diagnostic accuracy} = \frac{\text{Cumulative number of correct diagnoses}}{\text{Total number of diagnoses}} \quad (1)$$

As for examining the performance on interpretation, we employed accuracy, BERTScore [48], SentenceBert [49], and METEOR [50], which were widely used in related tasks, e.g., text summarization and generation [51, 52]. Concretely, the computation of interpretation accuracy is as follows:

$$\text{Interpretation accuracy} = \frac{\text{Cumulative number of correct interpretations}}{\text{Total number of interpretations}} \quad (2)$$

BERTScore [48] employs the BERT model [53] to determine the semantic similarity between reference and generated text, offering a context-aware evaluation of model performance. SentenceBert [49] measures sentence similarity using a BERT model that generates dense vector representations, facilitating efficient and accurate semantic comparisons. METEOR [50] assesses the harmonic mean of unigram precision and recall, utilizing stemmed forms and synonym equivalence.

## Design of human evaluation

We also evaluated the predicted DDx and interpretation with experts' efforts. For diagnosis, physicians assessed the consistency between predictions and ground-truth. If a predicted diagnosis was deemed a synonym or subtype of the ground-truth, it was marked as correct. For the generated interpretations, we followed related works [21, 54] and evaluated them across three dimensions: correctness, completeness, and usefulness. Correctness measures the medical accuracy of statements, completeness assesses the comprehensiveness of the interpretation based on symptom descriptions, and usefulness measures the overall value of the interpretation for diagnosis. Each note was reviewed by two physicians, with a third resolving any discrepancies in grading.

## Rare disease cases

We extracted cases of rare diseases for evaluation. We first determined the scope of rare diseases by following the definition from the National Organization for Rare Disorders (NORD)[4] and obtained a list of 1092 rare diseases. See the full list of rare diseases in Supplementary Data 10. Then, we extracted the cases with ground-truth diagnoses involving any rare disease in the list. Following this extraction guideline, we obtained 142 rare disease cases as a subset of the data. The extracted notes involving rare diseases are shown in Supplementary Data 11.

## Data availability

The constructed dataset is available in Supplementary Data 6.

---

[4] https://rarediseases.org/rare-diseases/

# Acknowledgments

This work was supported by the National Institutes of Health's National Center for Complementary and Integrative Health under grant number R01AT009457, National Institute on Aging under grant number R01AG078154, and National Cancer Institute under grant number R01CA287413. The content is solely the responsibility of the authors and does not represent the official views of the National Institutes of Health. We also acknowledge the support from the Center for Learning Health System Sciences.

# Author contributions

S.Z. and R.Z. designed the study. S.Z. contributed to the literature search and model evaluation. S.Z. and S.D. performed the data collection, prompt design, and model construction. S.Z., R.Z., and G.M. discussed and arranged data annotation and human evaluation. S.Z., M.L., and R.Z. conducted experimental design and manuscript drafting. R.Z. supervised the study. All authors contributed to the research discussion, manuscript revision, and approval of the manuscript for submission.

# Competing interests

The authors declare no competing interests.